\documentclass[letterpaper, 10 pt, conference]{ieeeconf}  

\IEEEoverridecommandlockouts                              

\usepackage{cite}
\usepackage{amsmath,amssymb,amsfonts}

\usepackage{graphicx}
\usepackage{textcomp}
\usepackage{xcolor}
\usepackage{hyperref}
\usepackage{svg}
\usepackage{multirow}
\usepackage{subcaption}
\usepackage{algorithm}
\usepackage{algpseudocode}
\usepackage{CJKutf8}
\usepackage{array}
\def\BibTeX{{\rm B\kern-.05em{\sc i\kern-.025em b}\kern-.08em
    T\kern-.1667em\lower.7ex\hbox{E}\kern-.125emX}}
\graphicspath{ {image/} }

\title{\LARGE \bf
Revolutionizing Battery Disassembly: The Design and Implementation of a Battery Disassembly Autonomous Mobile Manipulator Robot(BEAM-1)
}

\author{Yanlong Peng$^{1}$, Zhigang Wang$^{2}$, Yisheng Zhang$^{1}$, Shengmin Zhang$^{1}$, Nan Cai$^{3}$, Fan Wu$^{4}$, Ming Chen$^{1}$*
\thanks{This work was supported by the Ministry of Industry and Information Technology of China for financing this research within the program "2021 High Quality Development Project (TC210H02C)"}
\thanks{$^{1}$ School of Mechanical Engineering, Shanghai Jiao Tong University, Shanghai, China. (e-mail: \{me-pengyanlong, zys\_edward, zhangshengmin, mingchen\}@sjtu.edu.cn}%
\thanks{$^{2}$ Intel Labs China, Beijing, China. (e-mail: zhi.gang.wang@intel.com)}%
\thanks{$^{3}$ Faculty of Mechanical and Electrical Engineering, Kunming University of Science and Technology, Kunming, China. (e-mail: cainan2600@stu.kust.edu.cn)}%
\thanks{$^{4}$ Beijing-Dublin International College, Beijing University of Technology, Beijing, China. (e-mail: ericwufan@163.com)}%
}

\begin{document}

\maketitle
\thispagestyle{empty}
\pagestyle{empty}

\begin{abstract}
The efficient disassembly of end-of-life electric vehicle batteries(EOL-EVBs) is crucial for green manufacturing and sustainable development. The current pre-programmed disassembly conducted by the Autonomous Mobile Manipulator Robot(AMMR) struggles to meet the disassembly requirements in dynamic environments, complex scenarios, and unstructured processes.
In this paper, we propose a Battery Disassembly AMMR(BEAM-1) system based on NeuralSymbolic AI. 
It detects the environmental state by leveraging a combination of multi-sensors and neural predicates and then translates this information into a quasi-symbolic space. In real-time, it identifies the optimal sequence of action primitives through LLM-heuristic tree search, ensuring high-precision execution of these primitives. Additionally, it employs positional speculative sampling using intuitive networks and achieves the disassembly of various bolt types with a meticulously designed end-effector.
Importantly, BEAM-1 is a continuously learning embodied intelligence system capable of subjective reasoning like a human, and possessing intuition. A large number of real scene experiments have proved that it can autonomously perceive, decide, and execute to complete the continuous disassembly of bolts in multiple, multi-category, and complex situations, with a success rate of 98.78\%.
This research attempts to use NeuroSymbolic AI to give robots real autonomous reasoning, planning, and learning capabilities. BEAM-1 realizes the revolution of battery disassembly. Its framework can be easily ported to any robotic system to realize different application scenarios, which provides a ground-breaking idea for the design and implementation of future embodied intelligent robotic systems.

\end{abstract}

\section{INTRODUCTION}

The booming development of the worldwide electric vehicle industry \cite{bat1} has put forward new requirements for resource conservation, green manufacturing, and low-carbon sustainability. A large number of end-of-life electric vehicle batteries (EOL-EVBs) are in urgent need of disassembly and recycling. The present disassembly process for EOL-EVBs largely relies on manual disassembly lines to address challenges arising from different structures, deformations, and rust corrosion in EVBs. Repetitive tasks, notably the removal of bolts constituting approximately 40\% of all disassembly activities \cite{MR1}, result in significant labor costs.

In the field of intelligent disassembly, based on the combination of the intelligent vision system and robotic arm, the planning and modular semi-automatic disassembly can be realized \cite{bat2, bat3}. The development of customized cognitive robot systems can solve some of the disassembly work under ideal conditions \cite{batsys1, batsys2}. Other research has been done to accomplish specific types of bolt disassembly by aiding a priori knowledge through targeted designs \cite{MR1, MR2}. However, challenges such as multiple types, large scale, uncertainty in disassembly scenarios, and unstructured disassembly processes are difficult to be effectively and comprehensively addressed.

Mobile Manipulator Robot(MMR), which consists of a high-precision and compact manipulator and a freely movable chassis, provides new solution ideas for battery disassembly. Currently, depending on the application scenarios, MMR makes significant contributions in multiple fields \cite{1, 2, 3, 4}. Moreover, the Autonomous Mobile Manipulator Robot (AMMR) demonstrates higher robustness in tasks with static, dynamic, and even unknown environments in industrial scenarios \cite{AMMR1, AMMR2, AMMR3}. Unfortunately, current execution systems using end-to-end networks and high-precision sensors have difficulty in meeting the millimeter-level control accuracy required for battery disassembly \cite{endtoend}. The method based on deep reinforcement learning makes it even more difficult to accurately determine the right position for mobile disassembly \cite{Navigation}. Therefore, there is an urgent need for research on battery disassembly systems that can successfully carry out intricate, precise, and dynamic disassembly tasks with strong autonomous decision-making capabilities.

\begin{figure*}[th]
    \centering
    \includegraphics[scale=0.49]{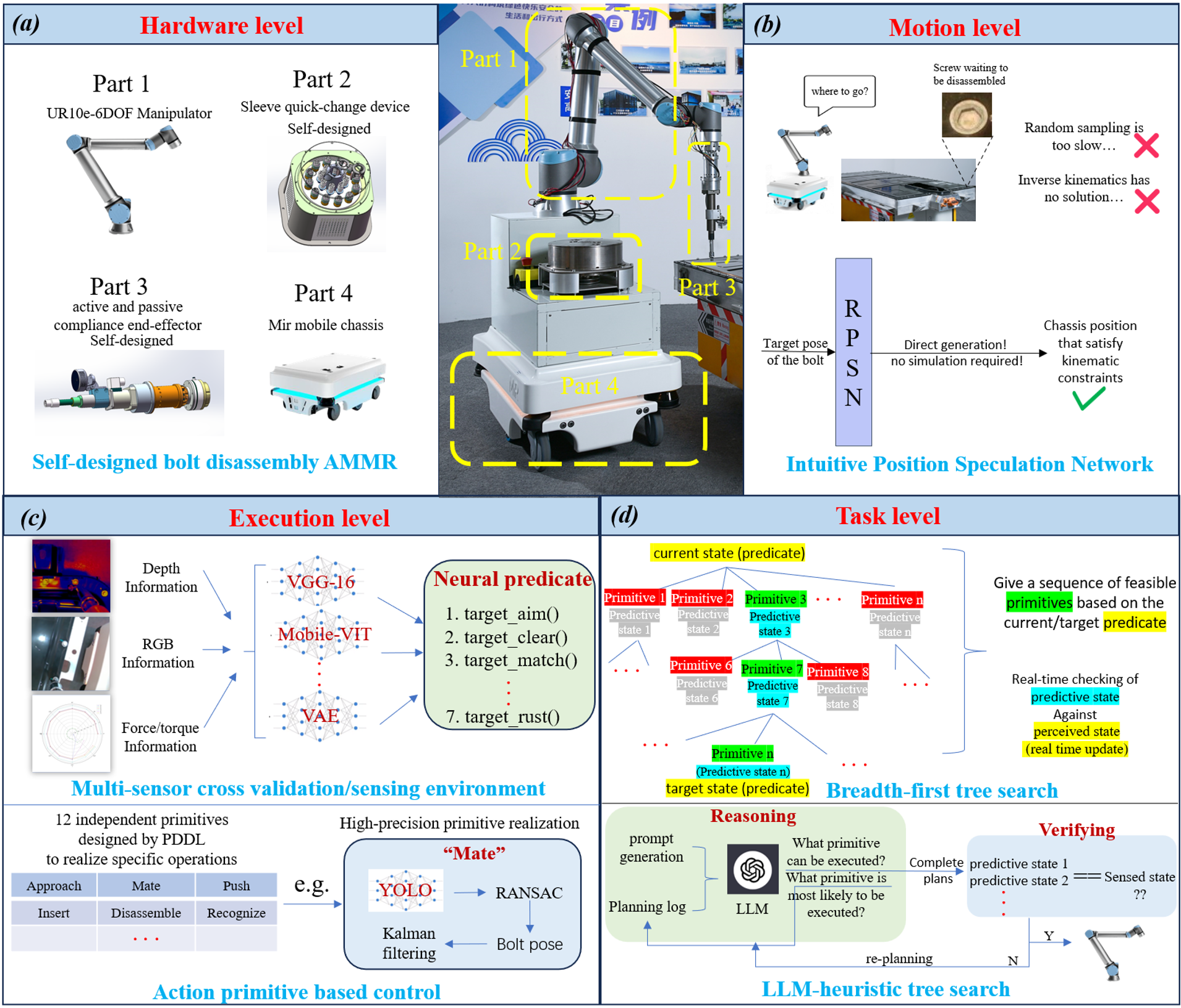}
    \caption{System architecture of our BEAM-1. According to different functions, it can be categorized into (a) body composition-Hardware level, (b) intuition-guided motion sampling algorithm-Motion level, (c) predicate and primitive-based high precision control-Execution level, and (d) LLM-heuristic tree-searching task planning-Task level.}
    \vspace{-3mm}
    \label{FIG:1}
\end{figure*}

Aiming at this difficulty, based on the NeuralSymbolic AI \cite{TeamWiththeStrong, NS-start3, shuangxitonglilun, NS-start1}, this paper designs a battery disassembly autonomous mobile manipulator robot system, BEAM-1, with autonomous perception, automatic planning, precise execution and continuous learning capability. (nickname BEAM, which is composed of 4 letters from Battery disassEmbly AMmr, that is, battery disassembly autonomous mobile manipulator robot). BEAM-1 possesses humanlike thinking ability, its subjective logic is ensured by symbolic reasoning and knowledge-driven logical framework, and its intuition and perception ability is generated by neural network and probability model. Without any human intervention, BEAM-1 can think in real-time according to the dynamic and complex environment it perceives, plan and execute the optimal sequence of action primitives, and efficiently complete the disassembly of multiple types of bolts of batteries. It is worth emphasizing that any primitive sequence of BEAM-1 is not achieved by enumerating pre-programming for task types.
Our BEAM-1 has the following five outstanding innovations:
\begin{enumerate}
\item Functional integration: Designed for the task requirements of EOL-EVBs, our BEAM-1 integrates a robotic arm, mobile chassis, active and passive compliance end-effector, and sleeve quick-change device, making it capable of accomplishing a wide range of tasks.
\item Task planning: Considering the unstructuredness and uncertainty of the disassembly scenario of used batteries, we design a task planning system based on NeuralSymbolic AI. The system utilizes neural predicates to dynamically perceive complex environments and convert them into symbolic representations. Then uses logical search to complete reasoning and plan the optimal sequence of action primitives.
\item Heuristic search: To improve the planning efficiency, we introduced a large language model(LLM) in the logical search and realized heuristic tree search, thus realizing efficient task planning.
\item Motion planning: To address the inefficiency of position speculation at the motion planning level, we introduce a Robot Position Speculation Network(RPSN). This allows BEAM-1 to quickly sample moving positions that satisfy kinematic constraints.
\item Continuous learning: In this embodied intelligence system, the neural predicates, LLM-based heuristic search, and RPSN can learn continuously based on the disassembly history. This allows the robot to continuously improve its intelligence.
\end{enumerate}

\section{Macro Architecture of BEAM-1}

As shown in Figure\ref{FIG:1}, the whole system includes four levels: hardware, execution, task, and motion. They are the indispensable parts that make up an AMMR, and also the core innovations covered by our solutions to the various challenges in battery disassembly.
The body of BEAM-1 consists of a 6-DOF collaborative robot, a mobile chassis, and a connecting structure. We designed an active and passive compliance motorized end-effector with multi-sensor integration and a quick-change device capable of replacing 20 different sleeves \cite{ZhangShengmin}.

BEAM-1 fuses multi-sensor information at the execution level, using pre-defined neural predicates to transform the environmental information captured by the sensors into a quasi-symbolic space. Meanwhile, the action of the robotic arm to complete the disassembly is decomposed into the minimal indivisible unit based on a priori knowledge, which is defined as the action primitive by using Planning Domain Define Language(PDDL) \cite{1998pddl, pddlstream}. Each action primitive has high-precision implementation logic and algorithms \cite{zhangyisheng}. The task level uses a logic tree search in symbol space to find the best sequence of primitives from the current state to the goal state \cite{zhanghengwei2}. We introduce LLM heuristic tree search to solve the problem of search space explosion and speed up the task planning search process \cite{zhangyisheng2}. At the motion level, the ability to speculatively sample based on intuition is given to BEAM-1 to improve its motion agility. We re-characterize the robot's forward and inverse kinematics using differentiable programming and incorporate them into ground truth-free guided neural network training, proposing RPSN \cite{rpsn}. The position of the target given by the RPSN satisfies the kinematic constraints without any simulation. The next chapter will provide a detailed exposition of the design principles and relevant technical details of BEAM-1 and articulate the greatest strength of the entire system, the ability to continuously learn.

\section{Core Technical Details}
\subsection{Hardware Design}

\begin{figure}[th]
    \centering
    \includegraphics[scale=0.24]{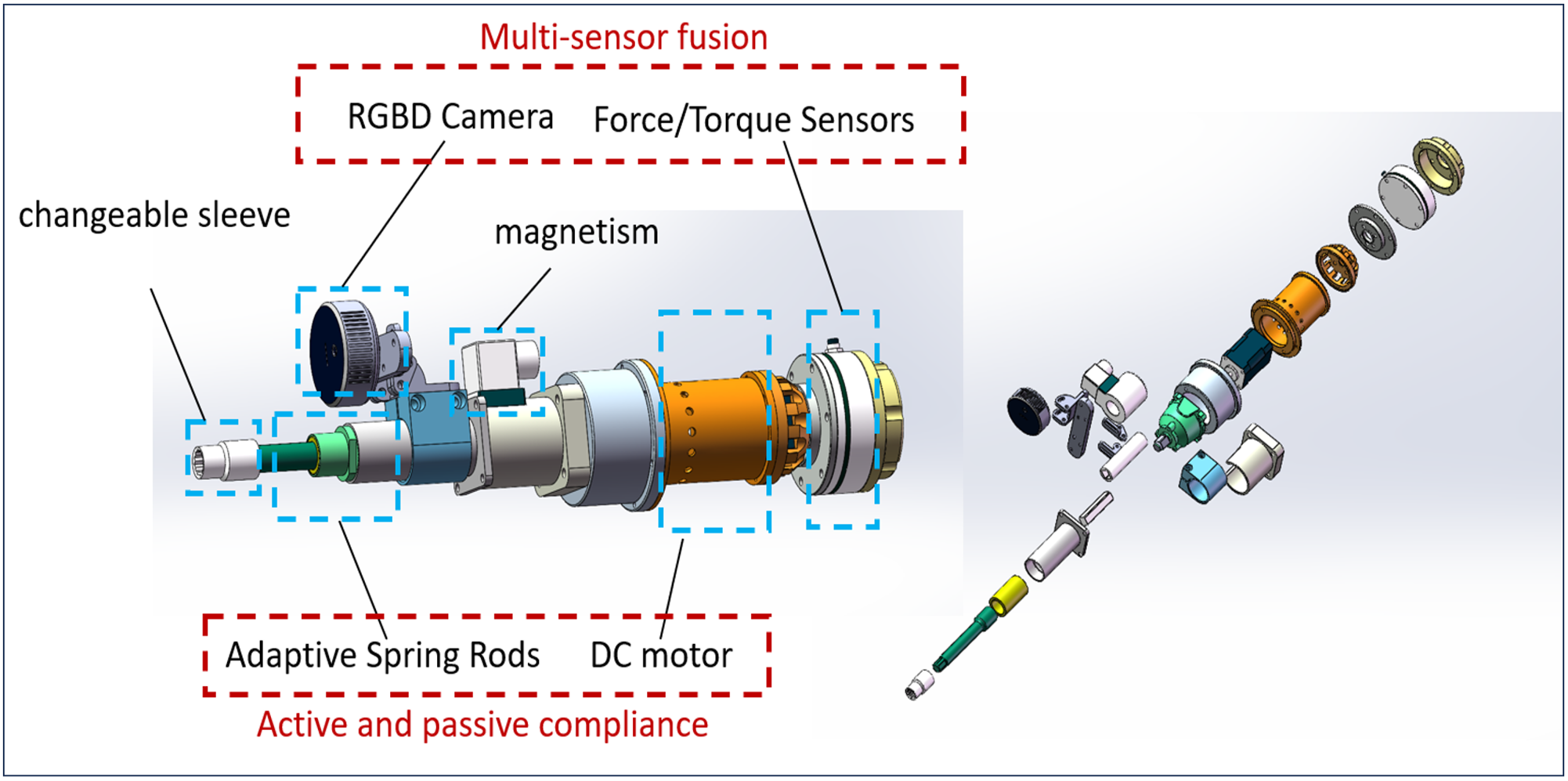}
    \caption{Active and passive compliance end-effector.}
    \vspace{-3mm}
    \label{FIG:2}
\end{figure}

A dexterous, stable, and agile actuator is essential to meet BEAM-1's disassembly requirements for multiple types of bolts. Our self-designed active and passive compliance electric torque end-effector with multi-sensors are shown in Figure\ref{FIG:2}. The real-time environmental information captured by the ATI multi-axis force/torque sensor and the Intel RealSense RGBD camera will be transformed into a quasi-symbolic representation of the current state via neural predicates. The DC motor will have different rotational speeds and directions in different primitives, providing active compliance for the disassembly primitive sequence and ensuring high-quality bolt engagement. Meanwhile, components such as springs, square shafts, and plain bearings transmit motor torque to the long rod fitted with a sleeve and give passive compliance to the entire end. The selective usage of the magnetic suction module aids in the disassembly of the magnetic bolts.

\begin{figure}[th]
    \centering
    \includegraphics[scale=0.24]{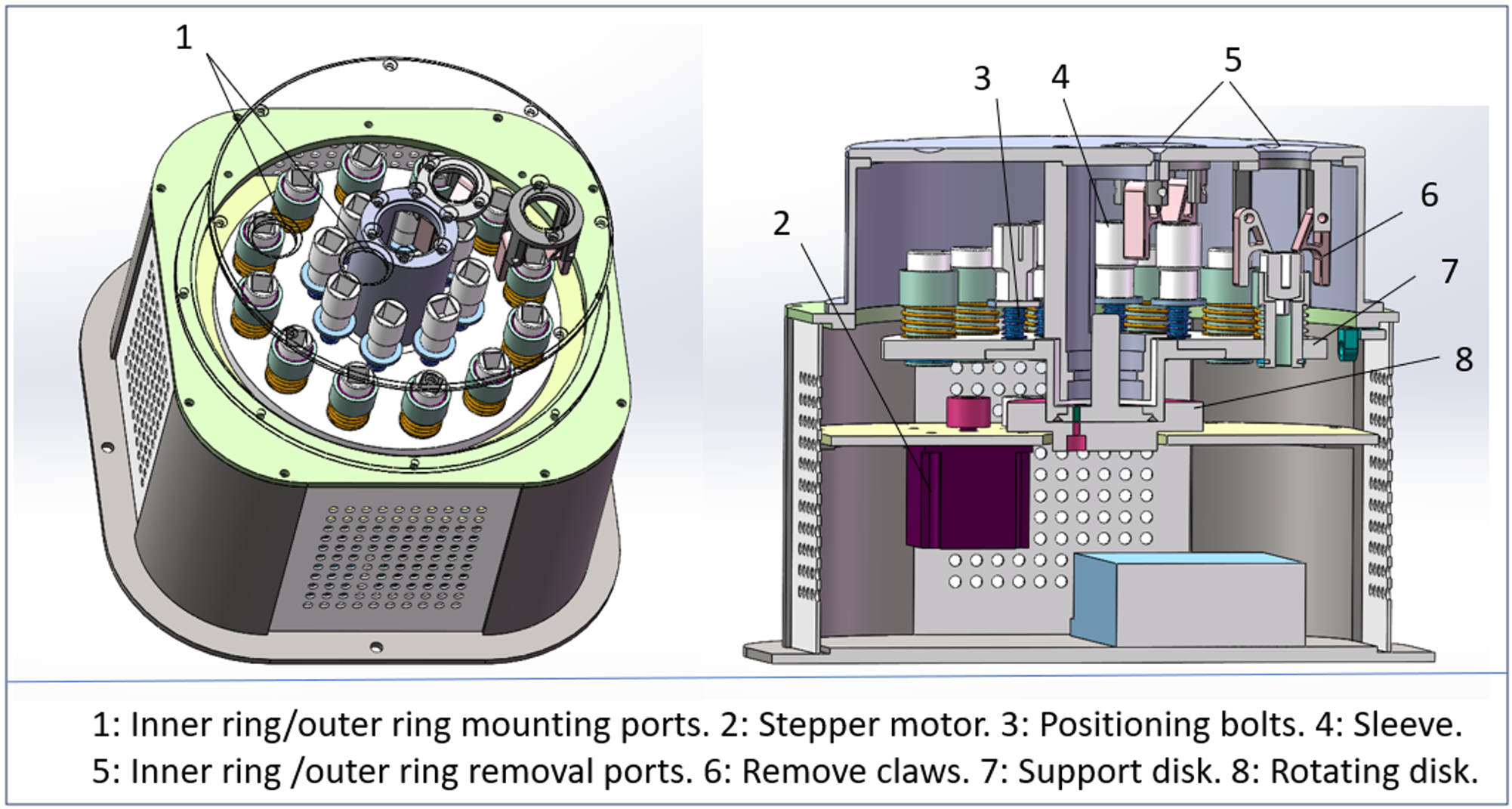}
    \caption{Quick-change disk with 20 different types of sleeves.}
    \vspace{-3mm}
    \label{FIG:3}
\end{figure}

In the fully autonomous disassembly task, when BEAM-1 realizes that the current actuator end sleeve does not match the bolt to be removed, its task planning process will involve primitive "Change". Therefore, we designed a stepper motor-controlled rotary quick-change disk as shown in Figure\ref{FIG:3}, with a choice of 20 different types and sizes of sleeves, which greatly expands the range of bolts that can be disassembled by this actuator. The robotic arm can remove the current sleeve at the removal port and install the new one at the mounting port utilizing a plugging operation.

\subsection{High Precision Control Based on Neural Predicates and Action Primitives}

Most current robotic systems heavily rely on high-precision sensors to perceive their environment. These systems execute predefined programs to perform corresponding robot operations under specific conditions, such as distance parameters. However, this approach fails to address the uncertainty during the battery disassembly process in highly dynamic environments. Different batteries, various bolts, and diverse disassembly scenarios cannot be universally dismantled using a standardized predefined method.

To achieve a more intelligent system, we introduce neural predicates to help BEAM-1 for environment state recognition based on the NeuralSymbolic AI. Each neural predicate can be regarded as a neural network, which maps the multi-sensor perception information of the environment to the quasi-symbolic space to complete the characterization of the state. As shown in Fig\ref{FIG:1}(c), we designed seven neural predicates such as target\_aim(). Real-time RGBD information and force/torque data are collected and input into pre-trained neural networks for processing. These networks provide probabilities for the corresponding states based on the input data. Neural predicates can be arbitrarily combined to describe the current complex state more accurately. The number of neural predicates can also be expanded to perceive more sophisticated environments. In our current work, we have achieved a 97.75\% success rate in recognizing the shape attribute of bolts using the VAE network, 100\% accuracy in recognizing the corrosion attribute of bolts, and 100\% accuracy in recognizing environmental states such as whether or not aligned by using the VGG network.

\begin{figure}[th]
    \centering
    \includegraphics[scale=0.30]{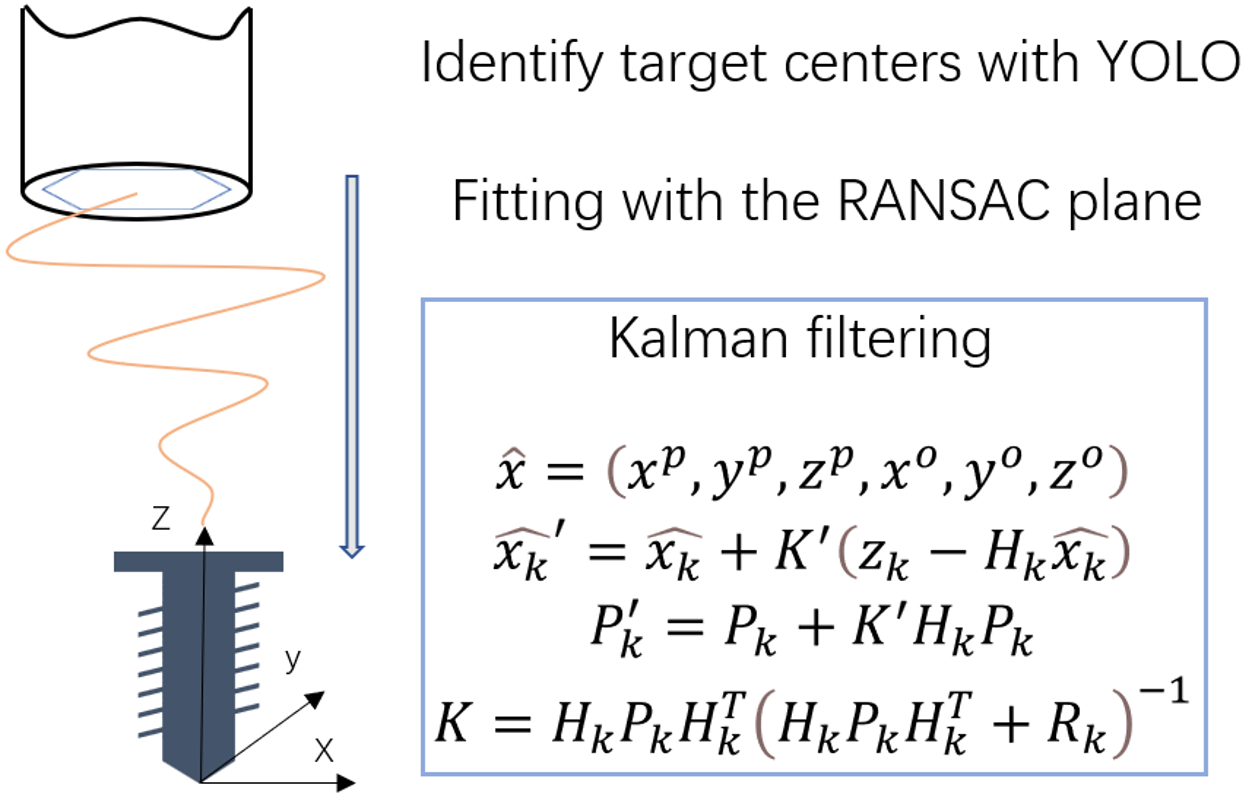}
    \caption{Implementation process of the primitive Mate: Kalman filtering and other algorithms to ensure the high accuracy of the primitive.}
    \vspace{-3mm}
    \label{FIG:4}
\end{figure}

Having accomplished the precise perception of the environment state, we realize high-precision control based on action primitives at the execution level. We subdivided the disassembly process and defined 12 action primitives such as Approach, Mate, Push, Insert, and so on. Each primitive is defined by PDDL with execution pre-requirements and execution target effects in symbol space, which will be used for searching during task planning.
The definition of primitives ensures that BEAM-1 can autonomously plan appropriate action sequences in dynamic and complex environments to cope with various environmental states and accomplish various tasks.
The accuracy of current popular control methods failed to meet the millimeter-level requirements in the disassembly environment \cite{endtoend, RT-2}, and this study uses manually implemented primitives to achieve high-precision accurate control while adding a layer of detection and verification at the primitive level.

As illustrated in Fig\ref{FIG:4}, When the BEAM-1 adjusts the end-effector to a coarse pose near the target bolt, the "Mate" primitive utilizes RGB information and YOLO to obtain bounding boxes and the target center. Subsequently, three-dimensional point clouds of adjacent planes are generated based on the bounding box and depth information. The normal vectors of adjacent planes are obtained using the RANSAC plane fitting algorithm and then transformed into the pose of the target bolt. Finally, the accuracy of the 6-DOF pose estimation is enhanced through the Kalman filtering algorithm. As the end effector progressively approaches the accurate position, the latest result becomes the final result when the covariance of process noise $\delta$ is less than the preset threshold $\epsilon$. Otherwise, BEAM-1 adjusts the end effector to the latest pose and repeats the aforementioned process. Experimental validation confirms that the pose estimation error of this system is sufficiently small(less than 0.5mm), ensuring a 100\% success rate in disassembly.

\subsection{Heuristic Search Strategy Based on Large Language Model}

\begin{figure*}[h]
    \centering
    \includegraphics[scale=0.42]{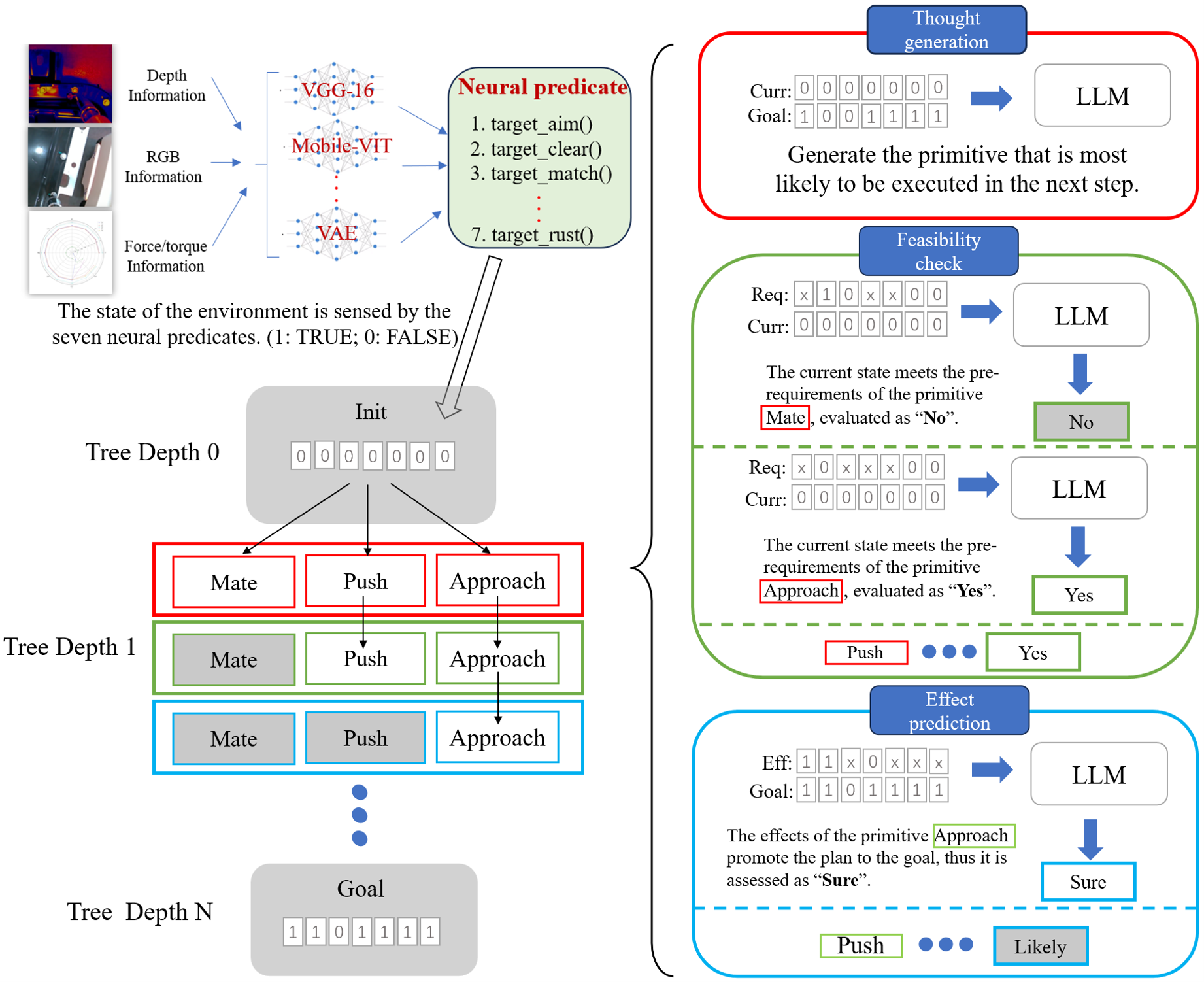}
    \caption{Reasoning engine with three prompting Sub-Engines based on LLM.}
    \vspace{-3mm}
    \label{FIG:5}
\end{figure*}

The unstructured battery disassembly process introduces more significant challenges, as we cannot predict the required disassembly primitive sequence in advance. To address this, we have developed a task planning system for BEAM-1 based on primitives and predicates.

At the task level, BEAM-1 generates the optimal sequence of action primitives from the current state to the goal state in the symbol space using the pre-requirements of the primitives and the current state given by the neural predicates, by using BFS tree search(Figure\ref{FIG:1}(d)).
In scenarios with complex tasks, the problem of exploding search space arises during the search for feasible solutions, leading to excessively long computation times. To tackle this issue, we introduce the LLM heuristic search, further enhancing the efficiency of task planning in unstructured environments.

The introduction of LLM divides the tree search process into two parts: the Reasoning Engine and the Verification Engine. The three prompting sub-engines of the Reasoning Engine greatly reduce the complexity of the original algorithm(Figure\ref{FIG:5}). Instructions and examples will be entered in advance to declare the task and standardize the output. After inputting the task goal into the Thought generation sub-engine, LLM will generate the next primitives that are most likely to be executed in that state based on the current state. The Feasibility checker sub-engine extracts the action primitives and related information directly from the output of the previous layer and then calls the LLM to determine whether the pre-requirements for the primitives are met. If it can be performed, the primitive is evaluated as "YES", otherwise as "NO". When the state does not satisfy the pre-requirements, searches along this primitive will be halted, effectively limiting the size of the tree by filtering the feasible actions. It is worth noting that such an approach allows for the feature of few-shot learning on prompt input. Finally, we developed the Effect prediction sub-engine, which predicts the state after completion of the primitive and compares it to the final goal state. This evaluates the effectiveness of each primitive. The effect of a primitive evaluated as "SURE" means that it is closer to the final goal than the result of an action evaluated as "LIKELY", and is more likely to be retained when choosing the optimal path. Upon completion of the inference, the Verification Engine uses the results of the primitive execution to predict the next moment state. By comparing with the perceived state of the next moment in real-time, the search will be repeated if it does not match.
Each layer of the tree performs such a process to achieve real-time autonomy for inference as well as validation, which in turn makes BEAM-1 excellent at handling multiple kinds of bolts and unstructured tasks.

We conducted three types of experiments to validate the sophistication and effectiveness of the task planning strategy. Firstly, we observed that the form of the descriptive language (keywords or narrative) does not decisively impact the experimental results within the same model. Secondly, through extensive ablation experiments, we found that the introduction of examples significantly improved the success rate of the algorithm during planning. Finally, in a comparative analysis of different search strategies, the planning success rate exceeded 95.7\%, far surpassing the unsupervised, individually primitive planning approach.

\subsection{Intuitively Guided Positional Speculative Sampling}

\begin{figure}[th]
    \centering
    \includegraphics[scale=0.26]{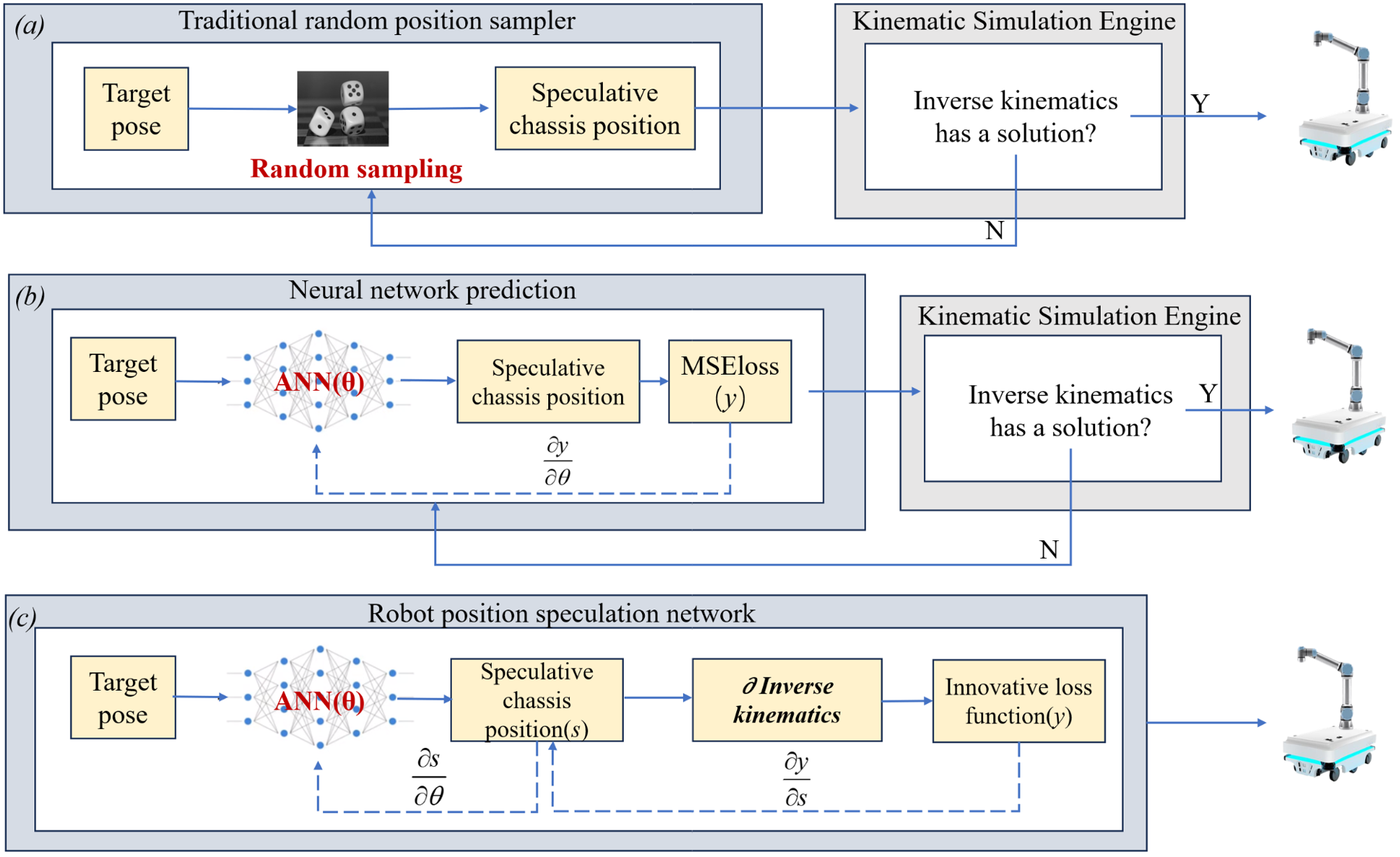}
    \caption{Comparison of three sampling methods for chassis position. (a) Widely used random sampling methods. (b) Training a network for position sampling by using a large amount of data. (c) The intuitive speculative sampling capability that BEAM-1 possesses. Add kinematics to computational graph in a differentiable way to participate in training.}
    \vspace{-3mm}
    \label{FIG:6}
\end{figure}
Dynamic scenarios with a wide range of bolt disassembly tasks require BEAM-1 to have the ability to sample the target position of the mobile chassis.

The target position should ensure that the robotic arm is within an appropriate range from the target bolt, and the manipulator's forward and inverse kinematics have a solution. Existing approaches(Figure\ref{FIG:6}(a)) use random sampling to determine where the chassis should move and rely on a kinematic simulation engine to verify whether it is kinematically solvable. This often requires multiple loop sampling simulations and uses significant computational resources. In addition, the deep learning approach (Figure\ref{FIG:6}(b)) requires a significant amount of data acquisition and ground truth labeling work. Once the bolts to be disassembled are on different packs in different locations, the change in dynamic scenarios means that the dataset needs to be recreated.

\begin{figure*}[th]
    \centering
    \includegraphics[scale=0.42]{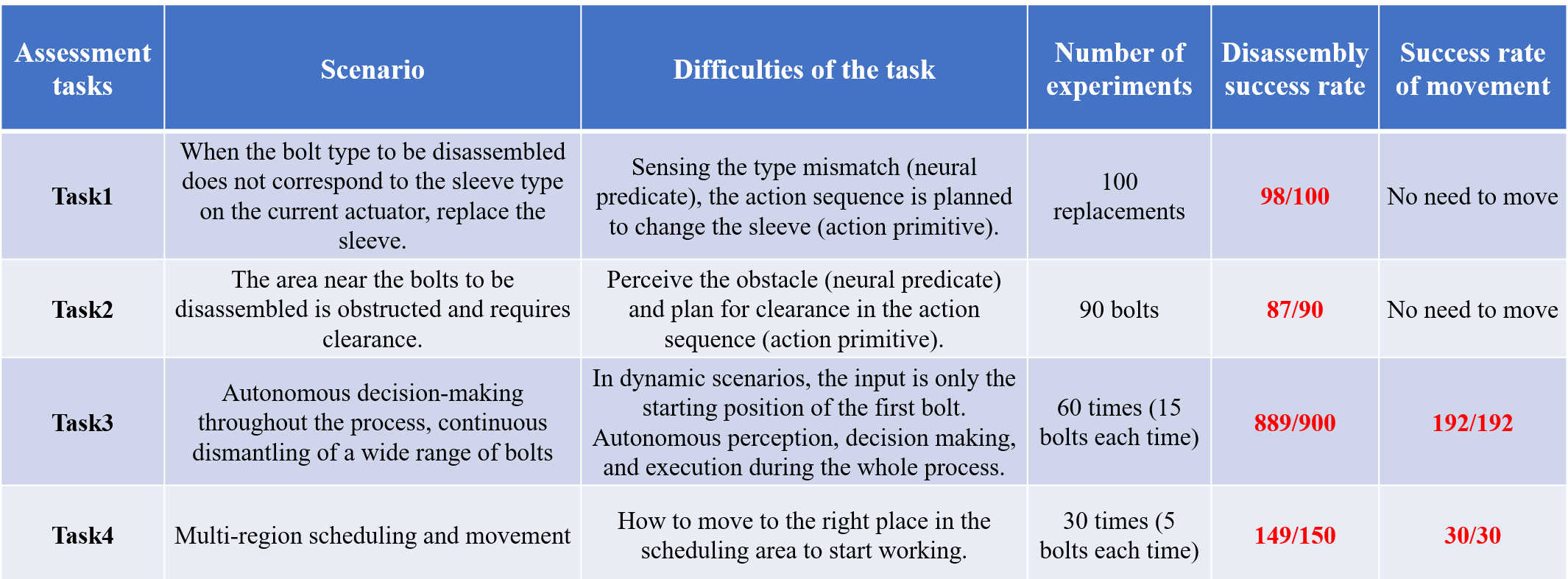}
    \caption{Details, execution difficulties, and experimental success rates of the four tasks in realistic scenarios.}
    \vspace{-3mm}
    \label{FIG:7}
\end{figure*}

Therefore, we equipped BEAM-1 with the RPSN as shown in Figure\ref{FIG:6}(c). We address the problems of function discontinuities and exception throwing during differentiable programming by adding the robot kinematics algorithmic constraints to the computational graph of ANN in a differentiable form. When movement is required, simply enter the pose of the target bolt, the RPSN can quickly give a high-confidence target position of the chassis where there must be a kinematic solution. It is worth emphasizing that the RPSN are trained without ground truth inputs. It relies on a unique form of a loss function to ensure that the anomalies of the differentiable engine become penalties to participate in backpropagation, which allows RPSN independent from specific scenarios and datasets thus able to truly have the ability to reason intuitively heuristically. Numerous experimental data indicate that randomized sampling and deep learning methods require an average of 31.04 times and an average of 3.06 times, respectively, to achieve point sampling that satisfies the kinematic constraints. Whereas the RPSN requires only 1.28 times, and the success rate of a single sample can be up to 96.67\%.

\subsection{Continuous Learning}

BEAM-1 can perform environment sensing through multi-sensors with neural predicates, autonomous decision-making through LLM heuristic tree search, position sampling through intuitive speculative networks, and efficient disassembly through high-precision primitives and mechanics. Thus, it exhibits fundamental characteristics of embodied intelligence. Furthermore, BEAM-1 also showcases continuous learning capabilities, as evidenced by:

\begin{enumerate}
\item Execution level: The recognition of states is dependent on neural networks. Therefore, continuous learning can be performed based on multi-sensor cross-validation as camera-guided and force-sensing combined. New neural predicates can be continuously expanded to enhance BEAM-1's perception of different environmental states.
\item Task level: The efficiency of heuristic search to find optimal primitive sequences can be continuously learned by utilizing LLM in-context learning.
\item Motion level: The existing RPSN itself is a trainable framework with a complete computational graph. Subsequently, additional physical constraints such as collision detection can be differentially programmed, ensuring the continuous expansion of the intuitive network.
\end{enumerate}

As a result, BEAM-1 can continuously learn new knowledge and skills from fresh data and experiences without forgetting previously acquired content. This capability enables it to adapt to various task scenarios, handle diverse sensory data, and address novel challenges. With embodied intelligence and continuous learning abilities, BEAM-1, akin to biological organisms, can achieve higher levels of intelligence and execution through its learning processes.

\section{Experiments and Results}
  
Aiming to validate the comprehensive performance under different task requirements and experimental scenarios of the BEAM-1 proposed in this work, we take the bolts on different EOL-EVBs as disassembled objects. Four tasks were designed for the experiments, including replacing the sleeve during disassembly, clearing the obstacles that affect disassembly, continuously disassembling a large number of screws, and disassembling tasks in designated areas, as shown in Figure\ref{FIG:7}. These tasks demand BEAM-1 to autonomously disassemble in dynamic, unpredictable, and complex environments without relying on any pre-programmed logic. This allows a comprehensive evaluation of its hardware design, perception effectiveness, control precision, planning quality, and motion intuition. Each task will emphasize different aspects, with specific execution challenges outlined in the figure. Particularly, in the case of large-scale scenarios where mobile disassembly is required, the movement success rate is also one of the metrics we need to focus on, which directly validates the effectiveness of intuitive sampling of the RPSN.

\begin{figure*}[th]
    \centering
    \includegraphics[scale=0.2]{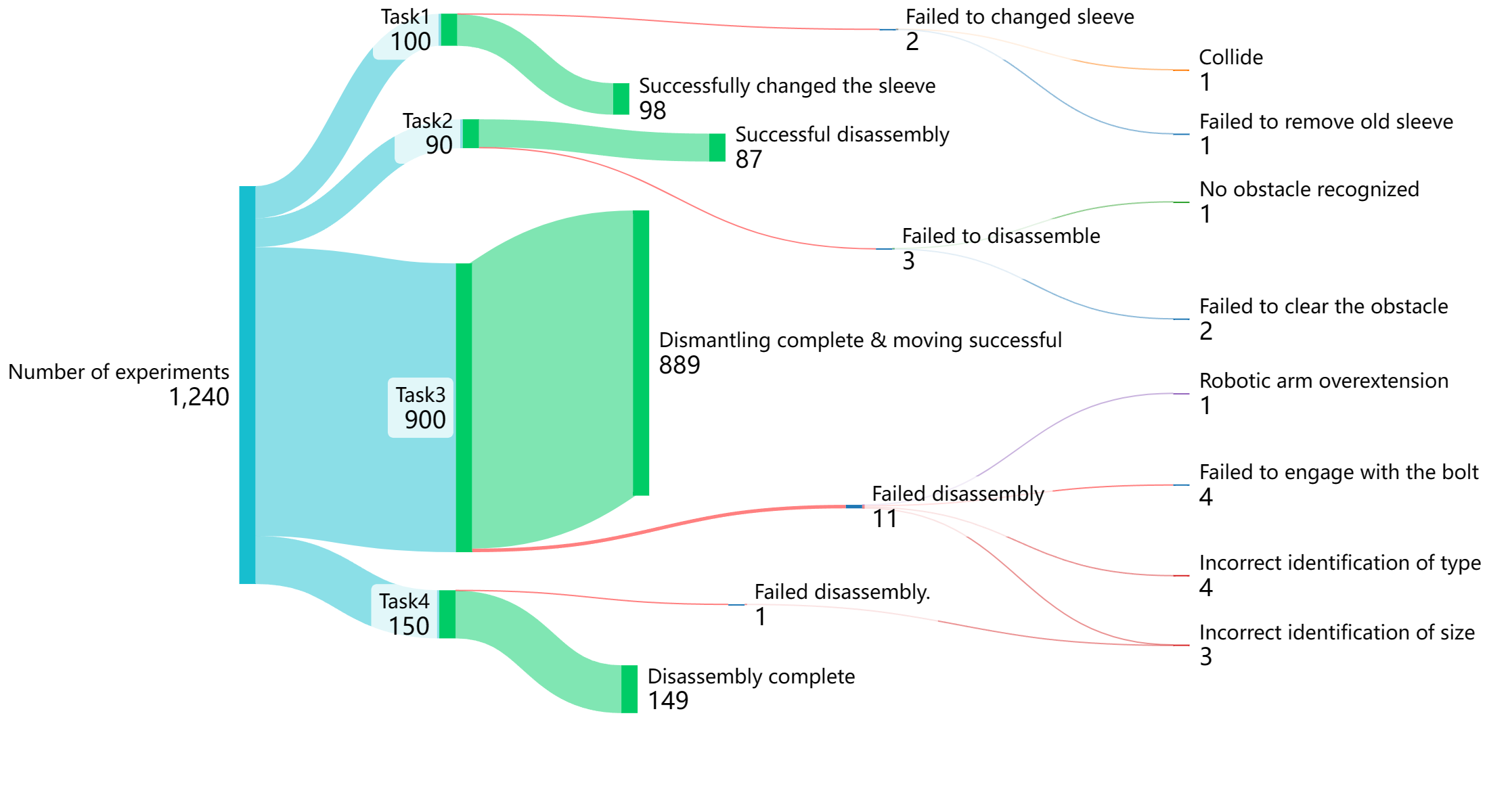}
    \caption{Experimental results and reasons of failure cases.}
    \vspace{-3mm}
    \label{FIG:8}
\end{figure*}

Figure\ref{FIG:8} illustrates the statistical data from multiple experiments involving the four different tasks, accompanied by a detailed record of specific reasons for some task failures. Based on the experiments, BEAM-1 demonstrated a 98\% success rate in detecting mismatches and completing sleeve replacements. The success rate for autonomously clearing obstacles and disassembling was 96.67\%. When faced with the disassembly of a large number of bolts, such as the continuous 15 bolts on one side of the battery pack in task 3, BEAM-1 only required the pose of the first bolt to autonomously plan and execute continuous disassembly, achieving an impressive success rate of 98.78\%. For scheduled tasks, the success rate of bolt disassembly in the designated area was 99.33\%. In all tasks involving autonomous movement, RPSN achieved a 100\% success rate in providing appropriate target sampling points with viable kinematic solutions.

The results show that BEAM-1 can accurately judge the environmental state through neural predicates, intuitively speculate the position through RPSN, generate the task plan in real-time through LLM-based heuristic search strategy,  and achieve high-precision disassembly control through carefully designed action primitives and end effectors. These capabilities together constitute the core competitiveness of BEAM-1 in the field of intelligent disassembly, enabling it to cope with various complex challenges.
A video demonstration of the disassembly tasks and more detailed information are available on the website.(see \href{https://sites.google.com/view/sjtubeam-1/home}{\textcolor{blue}{https://sites.google.com/view/sjtubeam-1/home}})

\section{CONCLUSIONS}
In this work, we build a highly autonomous and intelligent revolutionary disassembly system BEAM-1 for EOL-EVBs based on NeuralSymbolic AI. Compared with traditional pre-programmed or task-type-exhaustive robotic systems, BEAM-1 can autonomously carry out task and motion planning and complete continuous disassembly of multiple types of bolts in complex dynamic environments, with a success rate of 98.78\%. BEAM-1 is a continuously learning embodied intelligence system, endowed with the capacity for subjective reasoning akin to human and intuitive capabilities.
However, we have only applied it to the bolt disassembly scenario. In the future, we will fully utilize its learning ability to expand its working scenarios to disassemble fasteners, battery modules, battery cells, wire harnesses, and so on. 

This research aims to endow robots with genuine autonomous reasoning, planning, and learning capabilities through NeuroSymbolic AI. The universality of the NeuroSymbolic AI framework allows for its seamless application to diverse robotic systems, facilitating adaptation to various application scenarios. This innovative approach provides valuable insights for the design and implementation of future embodied intelligent robot systems.

\addtolength{\textheight}{-11cm}   




\section*{ACKNOWLEDGMENT}

The authors express their sincerest thanks to the Ministry of Industry and Information Technology of China for financing this research within the program "2021 High Quality Development Project (TC210H02C)".


\bibliographystyle{unsrt} 
\bibliography{ref}

\end{document}